\documentclass[lettersize,onecolumn, journal]{IEEEtran}
\usepackage{amsmath,amsfonts}
\usepackage{algorithmic}
\usepackage{algorithm}
\usepackage{array}
\usepackage[caption=false,font=normalsize,labelfont=sf,textfont=sf]{subfig}
\usepackage{textcomp}
\usepackage{stfloats}
\usepackage{url}
\usepackage{verbatim}
\usepackage{graphicx}
\usepackage{cite}
\usepackage{color}
\usepackage{colortbl}
\begin{document}

\title{Enhancing Visual Sentiment Analysis via Semiotic Isotopy-Guided Dataset Construction}

\author{
Marco Blanchini$^{1}$,
Giovanna Maria Dimitri$^{2}$,
Benedetta Tondi$^{1}$,
Tarcisio Lancioni$^{1}$,
Mauro Barni$^{1}$ \\[0.5em]
$^{1}$University of Siena, Siena (Italy)\\ 
$^{2}$University of Milan (Statale), Milan (Italy) \\

}

\maketitle

\begin{abstract}
Visual Sentiment Analysis (VSA) is a challenging task due to the vast diversity of emotionally salient images and the inherent difficulty of acquiring sufficient data to capture this variability comprehensively. Key obstacles include building large-scale VSA datasets and developing effective methodologies that enable algorithms to identify emotionally significant elements within an image.  These challenges are reflected in the limited generalization performance of VSA algorithms and models when trained and tested across different datasets. Starting from a pool of existing data collections, our approach enables the creation of a new larger dataset that not only contains a wider variety of images than the original ones, but also permits training new models with improved capability to focus on emotionally relevant combinations of image elements. This is achieved through the integration of the semiotic isotopy concept within the dataset creation process, providing deeper insights into the emotional content of images. Empirical evaluations show that models trained on a dataset generated with our method consistently outperform those trained on the original data collections, achieving superior generalization across major VSA benchmarks.

\end{abstract}


\section{Introduction}

Visual Sentiment Analysis (VSA) is a multidisciplinary field that combines computational methods with theoretical insights from psychology, semiotics, and affective science to understand and classify the emotional content of images.
VSA is applied in various fields including marketing \cite{VSOBorth2013}, social media analysis \cite{You2015}, artistic and cultural analysis of image creation and use \cite{Peng2016}, movie and media industry \cite{Campos2017}.

In healthcare, VSA contributes to the early diagnosis of mental disorders \cite{Zhou2020}, supports the curation of emotionally positive imagery to enhance hospital environments \cite{Campos2017}, and helps monitor patients’ emotional well-being.

VSA research requires the availability of high-quality image datasets labeled with the emotions aroused in the observers. The creation of such datasets faces the complex challenge of managing the vast range of visual element combinations that can evoke different emotions. In fact, the emotional impact of an image is influenced by numerous factors, including semantics, depicted subjects, colors, brightness, and many other features and their possible interactions.
This complexity makes it difficult to build a dataset covering all possible emotionally relevant visual combinations.
As a result, when such incomplete datasets are used to train modern VSA systems based on machine learning, some combinations may not be correctly recognized. Furthermore, the limited selection of images, sometimes focused on specific content like art or particular domains, introduces a bias that negatively affects VSA algorithms' ability to correctly classify images from different contexts.
Dataset biases often lead algorithms to focus on distracting elements that are not related to the emotional impact conveyed by a certain image. For instance, if within a dataset cars frequently appear in images evoking a positive sentiment, a model trained on such a dataset might erroneously associate all car images with positive emotions regardless of the context wherein they are used.

These limitations reduce the accuracy and generalization capability of machine learning algorithms, hindering their performance on different datasets. Such difficulties are clearly evident in the performance drop of models tested on datasets different from those used for training.

The most common approach to address these challenges is to construct increasingly larger datasets, based on the assumption that a greater number of images will naturally enhance the diversity of content and provide a broader range of emotionally evocative combinations. However, the construction of such large-scale datasets is incompatible with manual selection and annotation, hence requiring automated construction methods \cite{VSOBorth2013} \cite{t4s2017}. This in turn decreases the quality of emotional labels. In addition, most of the datasets built so far use heuristic approaches to increase diversity, without grounding the selection process on theoretically sound principles.
More recent research efforts have proposed alternative methodologies inspired by findings from cognitive science, aiming to guide automatic image selection and annotation through conceptually grounded principles \cite{VippSent_2025_WACV}. Models trained on datasets constructed using these techniques consistently outperform previous approaches in classification accuracy.

In this paper, we propose a new methodology, referred to as Selecta, for automatically constructing large, accurately annotated VSA datasets containing a wide variety of emotionally relevant patterns by leveraging multiple VSA models trained on a pool of pre-existing data collections to guide the selection and labeling of images crawled from the internet.

Selecta relies on two basic ideas: i) exploiting the availability of a pool of VSA models trained on different datasets to filter out images containing spurious or distracting patterns, and ii) leveraging the semiotic concept of isotopy to include in the dataset images with emotional patterns that were not present in the original collections.
Filtering is achieved by crawling the internet and retaining only images for which all, or the majority, of the pre-trained VSA models provide the same result. The inclusion of novel emotionally-relevant patterns unknown to the pre-trained models relies on the isotopy concept, which describes how elements present in a text or an image produce meaning 
In semiotics, isotopy refers to the repetition of semantically coherent elements that enables a text or image to communicate its message unambiguously to the observer. The central idea behind Selecta is to leverage this principle to construct an augmented large-scale dataset with reliable emotional labeling, thereby making it possible to train machine learning models focusing on genuinely relevant emotional features.

To validate our methodology, we focused on the classification of images into emotionally positive, negative and neutral classes, and used this approach to construct a new large dataset starting from an ensemble of VSA base models pre-trained on existing datasets, and comparing the performance of the original models with that of a model trained on Selecta\footnote{With a slight notation abuse in the following we use the term {\em Selecta} to indicate both the dataset and the methodology used to build it, the exact meaning being always clear from the context.}.
We then ran extensive experiments showing that the models trained on Selecta achieve better performance when tested on all major VSA datasets, outperforming both the algorithms trained on the original data collections and those trained on the most popular currently available datasets.

The rest of this work is organized as follows. In Section \ref{sec:State of the art}, we review the main VSA datasets available today. In Section \ref{sec:methodology}, we outline the theoretical foundations of Selecta and explain how they can be applied in practice. Section \ref{sec:simulation} introduces a set of simulations designed to demonstrate the effectiveness of Selecta in a simplified and controlled setting. In Section \ref{sec.actualdataset}, we describe the actual construction of Selecta dataset. In Section \ref{sec:experiment}, we describe the experiments we ran to validate our methodology. Finally, in Section \ref{sec:conclusion}, we draw our conclusions and outline directions for future work.

%-------------------------------------------------------------------------

%------------------------------------------------------------------------
\section{State of the art}
\label{sec:State of the art}

Developing a dataset for Visual Sentiment Analysis (VSA) involves significant challenges, including selecting reliable annotation methods and maintaining adequate quality and size \cite{Ortis2020}.
Existing datasets can be divided into two main categories: those designed to train Artificial Intelligence models and those intended for psychological studies on human emotional responses. 

The latter, which we refer to as PC datasets, rely on continuous classifications grounded in psycho-cognitive theories and organize emotions within the Dimensional Emotion Space (DES) \cite{dimensionalscherer2009}, where emotions are represented as points in continuous spaces defined by valence, arousal, and dominance, and are annotated in a highly controlled manner by experts in emotional psychology.
Although PC datasets provide highly accurate annotations, their limited number of images constrains their use for training AI models. The main examples are IAPS \cite{Iaps}, which includes 1,200 images divided into three classes (positive, negative, neutral); the IAPS-a variant, with 395 images based on Mikels’ eight emotions; and GAPED \cite{gaped}, which contains 730 images labeled as positive, negative, or neutral.

In this work, we focus on datasets designed to train VSA AI models. These kind of datasets requires a large number of images to cover the wide range of emotionally relevant possibilities and to ensure that the models are exposed to as much emotionally relevant information as possible during training. These datasets are mostly labeled with classes corresponding to discrete emotion categories, known as Categorical Emotion States (CES) \cite{dimensionalscherer2009}. The granularity of these classifications can vary, ranging from simple binary categories (e.g., positive vs. negative) to more varied sets, such as Paul Ekman’s six basic emotions \cite{Ekman1992}, or the eight primary emotions identified by Joseph Mikels \cite{mikels2005emotional}. Mikels' emotions can be easily mapped to positive or negative categories, while Ekman's emotions include surprise, which may encompass both positive and negative elements. Some datasets go beyond basic emotions by incorporating a broader set of categories \cite{emotic}.

One of the main challenges in building a VSA dataset for AI training, is finding the best compromise between highly accurate labeling and the necessity of annotating a large number of images.
To tackle this challenge, VSA datasets are usually created using three distinct labeling approaches: manual, hybrid, and automated.
Among the manually labeled datasets, we mention Emotion6 \cite{Emotion6}, which contains 1,980 images labeled by 15 annotators based on Ekman's emotional model, and FI \cite{FIYou2016BuildingAL}, including 23,308 images annotated by five individuals per image, also following Ekman's framework.  Another manually labeled dataset is Emotic, consisting of 18,360 images divided into 26 emotional categories, including the six basic emotions defined by Ekman and 20 additional emotional classes. 

A key limitation of these datasets is the use of uncontrolled online platforms for data collection, which introduces potential individual biases in emotion classification.
To mitigate this issue, some datasets adopt a hybrid approach consisting of an initial phase in which images are retrieved using emotionally salient queries, followed by a validation phase in which the emotional content is verified through interviews with human subjects. Examples of hybrid-labeled datasets include Flickr (90,139 images) \cite{katsurai2016} and Instagram (65,493 images) \cite{katsurai2016}. The images in these datasets are labeled according to three categories: positive, negative, and neutral emotions.
A further example of this category is Emoset \cite{yang2023L}, which includes 118,102 images labeled according to Mikels' eight emotions, with annotations provided by five individuals per image. The hybrid labeling approach adopted in \cite{yang2023L} led to improvements over manual labeling, both in the accuracy of models trained on this dataset and in their generalization capability on external data. Since Emoset  relies on interviews with multiple human subjects, the method lacks scalability.

Eventually, several large-scale datasets have been created by using fully automated approaches, which offer the advantage of enabling the annotation of vast amounts of data with minimal human intervention. Notable examples of this category include VSO \cite{VSOBorth2013} and T4S \cite{t4s2017}.
The Visual Sentiment Ontology (VSO) dataset comprises 500,000 images labeled with emotional adjective-noun pairs (ANPs) based on frequency of usage. Images were downloaded using queries composed of an emotion-related adjective and a commonly associated noun, and were labeled according to the emotional valence of the adjective. The dataset includes two classes: positive and negative emotions. While this approach enables the construction of large-scale datasets, it often lacks labeling accuracy \cite{VSOBorth2013}. T4S contains 1.5 million images sourced from Twitter, and annotated by analyzing the accompanying text.

Automatic dataset annotation significantly increases the volume of the dataset at the cost of lower labeling quality. A method to improve image labeling accuracy while using fully automatic annotation has been proposed in \cite{VippSent_2025_WACV} (VippSent dataset), by relying on the  application of techniques derived from cognitive science and semiotic studies. VippSent was built using an automatic labeling system based on the principles of visual semiotics and the mechanisms of emotional meaning in images. The dataset is divided into three classes: positive, negative, and neutral. VippSent was constructed by downloading images of the same subject associated with opposite and neutral adjectives, in order to encourage the classifiers to go beyond the subjects depicted by the images, to focus on the visual features that contribute to conveying emotions. The classes of emotionally charged images (both positive and negative) were enriched with artistic representations of emotional concepts.

In general, VSA algorithms exhibit limited cross-testing capabilities, with significant drops in accuracy when evaluated on datasets other than those used for training. Table \ref{Table1} presents the cross-dataset performance of models trained on the main VSA datasets.  
The table includes the main datasets among those that include both positive and negative classes, covering datasets with three classes, including a neutral class, as well as those based on Mikels' eight emotions \cite{mikels2005emotional}, which can be grouped into positive (Joy, Trust, Anticipation, Surprise) and negative (Sadness, Disgust, Anger, Fear) categories.  
This selection allowed us to perform cross-testing by training algorithms on each dataset and evaluating them on the others.  
To test algorithms trained on two-class datasets against three-class datasets, we removed the neutral class from the test set.  
Conversely, to test algorithms originally trained on three-class datasets against two-class datasets, we trained the algorithms using only two classes by excluding the neutral class.
As shown in Table~\ref{Table1}, the best performances in terms of generalization capability are achieved by models trained on sufficiently large datasets with accurate labeling techniques. This suggests that the limitations in the generalization capability are determined by an insufficient variety of emotionally relevant images contained in the training datasets or by unreliable labeling.

\section{Methodology}
\label{sec:methodology}

In this section, we describe the Selecta methodology for the construction of VSA datasets. We start by introducing the semiotic principles that Selecta relies on, then we explain how such principles can be used to construct a VSA dataset with enhanced diversity and generality. Eventually, we provide some insights into the way the construction of Selecta helps to reduce bias and to improve the generalization capability of the dataset.

\begin{table}[h!]
\centering
\caption{Cross-dataset accuracies of CLIP ViT-B/32 models trained on different datasets. Columns indicate the training dataset, rows the test sets.}
\renewcommand{\arraystretch}{1.5} 
\label{Table1}
\large % <-- prova anche \footnotesize o \scriptsize se vuoi ancora più piccolo
\resizebox{0.5\textwidth}{!}{%
\begin{tabular}{|l|c|c|c|c|c|c|c|}
\hline
\textbf{Test set} & \textbf{Emoset} & \textbf{Flickr} & \textbf{Insta.} & \textbf{Vipp} & \textbf{Vso} & \textbf{T4S} & \textbf{FI} \\
\hline
\textbf{Emoset}   & 96.04\% & 81.58\% & 72.25\% & 90.06\% & 65.75\% & 79.38\% & 86.30\% \\
\hline
\textbf{Flickr}   & 84.54\% & 75.19\% & 69.03\% & 62.22\% & 67.83\% & 25.05\% & 85.43\% \\
\hline
\textbf{Insta.} & 88.08\% & 74.36\% & 68.51\% & 57.85\% & 78.19\% & 25.50\% & 86.27\% \\
\hline
\textbf{Vipp}     & 91.12\% & 71.22\% & 56.70\% & 88.64\% & 86.50\% & 60.45\% & 84.78\% \\
\hline
\textbf{Vso}      & 65.74\% & 62.05\% & 67.99\% & 62.05\% & 73.26\% & 60.79\% & 61.24\% \\
\hline
\textbf{T4S}      & 61.21\% & 43.49\% & 39.27\% & 44.39\% & 56.99\% & 57.27\% & 60.41\% \\
\hline
\textbf{FI}       & 86.00\% & 79.19\% & 86.05\% & 88.15\% & 75.36\% & 82.70\% & 92.44\% \\
\hline
\end{tabular}%
}
\end{table}

\subsection{Isotopy in semiotic theory}

The isotopy concept, introduced by A. J. Greimas \cite{greimas1987meaning}, explains how meanings emerge from a text or an image, and it is also applicable to emotional meanings \cite{greimas1993semiotics}. Isotopies are semantic patterns created by the recurrence of terms or visual elements that converge toward a shared meaning. Their interplay ensures coherence and allows the text or image to convey specific emotional or semantic significance.

In the semiotic analysis of a text, the basic building blocks are 
\emph{lexemes}, words with autonomous meaning, such as \emph{dog}, 
\emph{cat} or \emph{man}. Lexemes can be attributed with \emph{semes}, 
which are semantic properties that characterize them. For example, the 
lexeme \emph{dog} can be associated with semes like \emph{animal}, 
\emph{quadruped}, or \emph{mammal}.  When a lexeme is placed in a context and combined with other lexemes,
additional semes can emerge from their interaction. For instance, in the expression dog barks,
semes such as anger, fear, or communication are not strictly inherent to the lexemes
but can be activated in context through their combination. If we expand the expression to
dog barks at the wind, further semes may appear, such as futility or frustration,
which arise from the interplay of lexemes within the text.
\emph{Isotopy} refers to a phenomenon where different expressions in a text 
share the same semes, thereby creating a homogeneous meaning.
For example, in the sentence: ``\emph{The dog barked, he slammed the door, 
and shouted},'' the expressions \emph{dog barked}, \emph{slammed the door}, 
and \emph{shouted} share the seme \emph{anger}. This recurrence generates 
an isotopy of anger, providing emotional coherence to the sentence.
Notably, the lexeme \emph{anger} does not appear explicitly, but the 
meaning emerges from the homogeneity of the semes in the text.

Isotopy was initially studied with reference to written texts but has since been extended to images through visual semiotics.
In this way, isotopy can also explain how meaning emerges from images. For example, in an image depicting a sunset on a tropical beach and a man drinking a cocktail, we can recognize semes such as {\em relax} and {\em vacation}. However, the individual visual lexemes, such as the sun, the sky, the beach, or the cocktail, evoke these semes only partially or indirectly. For instance, in a different context, such as during a storm, the beach would not evoke {\em relax} or {\em vacation} at all. The overall meaning emerges from the interaction and coherence among the various visual elements, just as it does in written texts \cite{floch2000visual}.

The principle of isotopy is activated whenever there is an intention to convey a message, whether through verbal language or visual means. In most cases, this process occurs unconsciously and concerns the compositional choices made by the enunciator in constructing the textual or visual message \cite{greimas1987meaning}.
When generating a message by taking a photograph, for instance, scenes whose elements are coherent with one another are chosen in order to converge toward the intended meaning \cite{sontag1977}. In the case of positively connoted events, specific colors and lighting conditions tend to be favored; conversely, negative events are typically represented through different color palettes and visual solutions \cite{boltanski1993souffrance}.

Isotopic mechanisms are recurrent in the vast majority of images produced and disseminated in communicative contexts, such as digital platforms, and can be leveraged to improve the effectiveness of algorithmic-based classification systems. In fact, isotopy can help explain the difficulty that models trained on one dataset face when classifying images from another dataset. The fact that emotions can be determined by the convergence of multiple elements, rather than by the mere presence or absence of individual ones, makes it harder to classify images coming from different contexts. On the other hand, understanding the isotopy mechanism can help improve the capabilities of VSA algorithms.

\subsection{Exploitation of isotopy for automatic dataset construction}

In this work, we rely on the isotopy principle to address the limited generalization capability of VSA models trained on existing datasets. 
 
The isotopy principle suggests that the emotional meaning of an image emerges from the combination of elements and their semantic value, which together convey meaning to the viewer. In contrast, AI models tend to rely on the presence of individual visual elements to classify images into different categories, which limits their ability to classify images based on abstract concepts such as emotions \cite{geirhos2019imagenet}\cite{zhang2019interpreting}.

Our hypothesis is that VSA models tend to exhibit a bias in emotion classification caused by the distribution of visual elements in the training data. Emotions are abstract concepts that cannot be expressed through unique and easily detectable visual elements, but rather through a multitude of possible combinations. A concrete lexeme such as "dog" corresponds visually into a single visual lexeme, characterized by clearly shared elements: muzzle, tail, legs. An abstract lexeme such as "love" can have a variety of visual representations, such as two lovers kissing or a heart symbol, which emerge not from individual elements alone but from specific contextual combinations. This broad and heterogeneous variety of visual forms used to represent abstract concepts such as emotions is at the root of classification biases.
If in a given dataset certain emotional classes contain images with recurring elements, such as specific colors or subjects, the model may associate those elements with that emotion, even when they appear in images with a different emotional value, thus producing a classification bias. The type of images present in different datasets influences the presence of such biases: for instance, artistic photography tends to associate certain colors or subjects with particular emotions \cite{machajdik2010affective}, whereas social media shows a more heterogeneous distribution. On social media, many everyday situations are represented only in emotionally positive images, while their negative representations are underrepresented \cite{masciantonio2025positivity}. Taken together, these biases limit the generalization capability of models across datasets different from those used in training.

To mitigate such biases, it is essential to build large and diverse datasets in which different elements appear across various emotional classes, so that the model learns to recognize emotions from complex patterns that genuinely evoke them, rather than from recurring individual visual elements. It is also necessary to ensure reliable automatic image labeling, in order to expand the dataset without introducing bias or noise.

\subsection{Pipeline of Selecta}

\begin{figure}
    \centering
    \includegraphics[width=\linewidth]{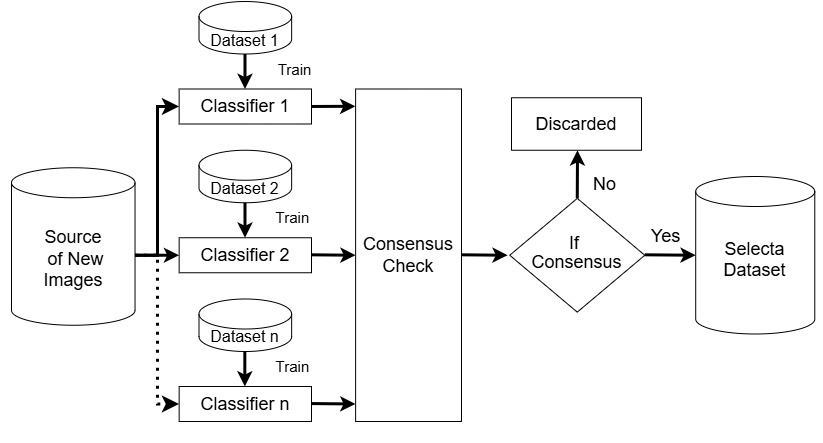}
    \caption{Overview of Selecta dataset construction methodology. }
    \label{fig:selection_process}
    \vspace{0.5em}
\end{figure}

Figure \ref{fig:selection_process} illustrates the procedure we are proposing to build a coherent and balanced emotional dataset based on the principle of isotopy. The process unfolds as follows.
We assume to have access to a large source of unlabeled images, as well as to $n$ labeled data collections that offer a limited and potentially incomplete characterization of images conveying emotions belonging to predefined classes (positive, negative, and neutral emotions in our experiments). 
As a first step, the original datasets are used to train an ensemble of $n$ base VSA models. These models are then employed to independently annotate the unlabeled image collection.
The resulting independent labels are compared to assess whether a consensus exists among the base classifiers. Different consensus strategies can be applied. For small values of $n$, unanimous agreement may be appropriate, whereas for larger $n$, a predefined majority might suffice.
Only images for which a consensus is reached are retained to form the new dataset. Their labels are automatically determined by the agreed-upon classification from the ensemble of base models.

In building the dataset and conducting the simulations, we chose to use three datasets for training the algorithms employed in image selection, specifically three datasets that classify images into three emotion classes: positive, negative, and neutral, a categorization widely used in psychology \cite{Russell1980}. However, in principle, our methodology is applicable to any number of datasets with any number of emotion classes, provided that the images in the different datasets are classified according to the same classes.

The filtering and labeling methodology described in Figure \ref{fig:selection_process} leverages two fundamental principles of isotopy. The first principle is that the emotional meaning of an image arises from the coherence among multiple elements belonging to the same isotopy \cite{greimas1987meaning, floch2000visual}, rather than from the presence of a single element. Our filtering system increases the likelihood of selecting images that contain concordant elements capable of evoking a specific emotion, while simultaneously reducing the influence of biases linked to individual visual components.

The second principle is that elements within an isotopy tend to recur in regular combinations across images expressing the same emotional polarity. This pattern emerges because image creators - either consciously or not \cite{sontag1977} - tend to favor compositions that include as many reinforcing elements as possible to support the intended emotional message. As a result, statistically, if an image contains certain elements that form an emotional isotopy, it is highly probable that it also includes others that are consistent with it.
Consequently, images that are classified consistently by multiple classifiers are not only likely to contain a coherent emotional isotopy, but also contribute new, representative elements of the same isotopic chain, that were not present in the original datasets, thereby enriching the resulting dataset.

\subsection{Isotopy-based analysis}
\label{subsec.patterns}

In this section, we provide some insights about the mechanism whereby Selecta contributes to reducing bias and improving the generalization capability of the dataset.

It is well established that deep learning models classify images by identifying specific visual patterns learned during training \cite{geirhos2019imagenet, zhang2019interpreting}. We hypothesize that these patterns can be roughly grouped into three categories based on their relationship with the emotional content of the image they belong to.

\paragraph{Pattern A} Emotionally relevant visual patterns, that is, sets of visual lexemes that, when combined within an image, generate an overall meaning (sememe) that is part of an emotional isotopy and thus evokes a specific emotion in the viewer.
For example, combining the visual lexeme of a house with that of fire coming out of a window can evoke concepts such as destruction and danger, conveying a negative emotional value to the image.
\paragraph{Pattern B} These are visual lexemes that, although appearing frequently in a specific emotional class, do not generate a typical emotional effect when taken in isolation.
When combined with other coherent elements (Pattern A), however, they can contribute to conveying a clear emotional meaning. If during training these patterns appear only within a single emotional class, they may result in a classification bias.
For example, a skull is a visual lexeme commonly found in images labeled as negative. When paired with other elements such as war scenes, weapons, or destruction, it contributes to the formation of a negative isotopy. However, a skull can also appear in positive contexts - for instance, a child dressed up for Halloween.
If a model has been trained to associate the skull solely with negativity, it may induce a bias. In such cases, the classifier might overlook other positive cues in the image - such as the festive atmosphere created by the child’s smile or the presence of candies - and incorrectly classify the image as one conveying a negative emotion.

\paragraph{Pattern C} These are visual lexemes that appear incidentally within images of a given emotional class but do not inherently carry an emotional meaning.
For example, if in a dataset chairs happen to occur only in positive images, the model may incorrectly learn to associate the presence of a chair with a positive emotion, leading to a classification bias.

The Selecta filtering procedure, based on consensus among models trained on different datasets, leverages the isotopy principle to enrich the dataset with Pattern A elements. In fact, images that receive consistent classifications across the base models are more likely to contain type A patterns, as it is unlikely that type B - and, even more, type C - patterns would lead to an agreement among the base classifiers. As more images are processed, type B elements tend to distribute more evenly across emotional classes, often balanced by stronger type A patterns. This reduces their biasing effect and enhances generalization. Meanwhile, the influence of type C patterns is minimized, as random elements are unlikely to appear consistently within a single class across different datasets.

Lastly, images containing type A patterns often include additional visual elements that reinforce the dominant isotopy, thereby enhancing the emotional richness and semantic depth of Selecta dataset. This enrichment is enabled by the statistically recurrent combinations that characterize elements within an emotional isotopy.

In the rest of the paper, we exploit the concepts discussed in this section for the construction of a three-class VSA dataset containing images conveying positive, negative, and neutral emotions.

\section{Simulations}
\label{sec:simulation}

To validate our methodology, we conducted some simulations to observe the behavior of a classifier based on Selecta dataset  within a simplified environment. Thanks to this simplified model, we are able to simulate mechanisms analogous to isotopy, and verify that the consensus-based filtering strategy can effectively leverage isotopic structures to construct an augmented dataset and train a classifier with improved generalization capabilities.

To start with, an image (and its content) is abstractly modeled as a multiset of symbols drawn from a basic alphabet, where 
a multiset (also known as a {\em bag}) is a non-ordered collection of symbols. In contrast to ordinary sets, in a multiset elements can appear multiple times, and the number of occurrences (called the multiplicity) matters. In our simulations, symbols represent a simplified version of the visual lexemes that make up an image. For example, each symbol can correspond to a meaningful shape, such as the silhouette of a "dog", or to a color, a certain brightness level, or any other visual element contributing to the overall meaning of the image.
More formally, given an alphabet $\mathcal{S} = \{s_1, s_2 \dots s_m\}$, an image $X$ is modeled as a non-ordered bag of symbols drawn from $\mathcal{S}$:
\[
    X  = \{x_1, x_2 \dots x_{k_X}\}, x_i \in {\mathcal S},
\]
where the number of symbols contained in $X$, $k_X$, may vary from one image to the other.

In semiotics, lexemes possess a form, given by the visual elements that compose them, and a semantic content, that is, the meaning they evoke in the viewer \cite{barthes1964}.
In our multiset model, each symbol is assigned an emotional polarity: positive, negative, or neutral.
The overall emotional label of the image is determined based on the emotional categories of the symbols contained in it.
If neutral symbols constitute an absolute majority, the image is classified as neutral ($nt$). Otherwise, it is classified as positive ($p$) if positive symbols outnumber negative ones, and negative ($n$) if the opposite is true. In case of a tie between positive and negative symbols, the negative label prevails. Formally:
\begin{equation}
   \ell(X) = \left\{
\begin{array}{ll}
nt & \text{if } \#nt > \frac{k_X}{2} \\
p & \text{if } \#nt \le \frac{k_X}{2}  ~ \cap ~ \#p > \#n  \\
n & \text{otherwise}
\end{array}
\right.
\label{eq:majority_rule}
\end{equation}

This mechanism provides a simplified representation of the tendency of emotional stimuli to prevail over neutral ones, and of negative stimuli to prevail over positive ones \cite{vuilleumier2005how}. 

The simplified representation of an image as a multiset of symbols is consistent with how semiotics analyzes image signification. Visual semiotics examines the meaning of images through the grammar of visual language, where visual elements play the role of words whose combination produces the overall sense of the image \cite{GrammaticaVisivakress2006reading}. In this context, meaning, including the emotional component, emerges from the overall visual context rather than from individual elements \cite{barthes1964}.
A limitation of the simplified image model used in our simulations is that the spatial position of elements is not taken into account, even though it can be crucial in some cases. However, according to the principle of isotopy, it is rare for the emotional impact of an image to depend solely on the arrangement of objects, as other coherent elements are often present that reinforce the overall emotional effect.

We also adopted a simplified mechanism to simulate a classifier trained on a dataset of images  (multisets in our case). Specifically, we implemented our classifiers based on the Intersection over Union (IoU) similarity metric, adapted to work on multisets. Given two images $X$ and $Y$, we define their Intersection over Union (IoU) similarity as:
\[
\text{IoU}(X, Y) = \frac{|\text{set}(X) \cap \text{set}(Y)|}{|\text{set}(X) \cup \text{set}(Y)|},
\]
where $\text{set}(X)$ and $\text{set}(Y)$ denote the sets of unique symbols contained, respectively, in the multisets $X$ and $Y$, and $|\cdot |$ indicates the cardinality of a set.

With the above definitions, let ${\mathcal X}_{tr} = \{X_1 \dots X_n\}$ be a set of images, and $\phi_{{\mathcal X}_{tr}}$ a base classifier based on ${\mathcal X}_{tr}$. Given a test image $Y$, the classifier computes IoU$(Y, X_i)$ for all the $X_i$'s in ${\mathcal X}_{tr}$ and assigns to $Y$ the same label of the most similar image in ${\mathcal X}_{tr}$. Ties are solved by looking at the second most similar image and so on.
The construction of Selecta dataset, then, proceeds as illustrated in 
Figure \ref{fig:selection_process}.

The goal of the simulations is to observe in a simplified and controlled environment how the Selecta dataset emerges from the original datasets ${\mathcal X}_i$'s, how the presence of isotopies influences the composition of Selecta dataset, and to what extent a classifier based on Selecta improves the performance of the base classifiers relying on the original datasets ${\mathcal X}_i$'s. 

In particular, we aim to verify two fundamental hypotheses. The first hypothesis is that Selecta effectively enforces the selection of images containing well-defined isotopies (Pattern A in Section \ref{subsec.patterns}), while reducing the impact of individual visual elements that may introduce classification bias (Patterns B and C in Section \ref{subsec.patterns}).
The second hypothesis concerns the ability of the Selecta to exploit the recurrence of elements that characterize images belonging to the same emotional class, as hypothesized by the principle of isotopy. In a context where the isotopy principle holds, Selecta should enable the selection of images containing new visually and emotionally relevant elements within the constructed dataset and increase the generalization capacity of algorithms trained on it.

\subsection{First simulation}

The first simulation is a very simple one. We considered a case in which the number of original datasets used to build the original classifiers is equal to 3.
Given the limited number of classifiers, we adopted a simple consensus mechanism, whereby Selecta retains only the images for which all three classifiers produce the same result.
We ran the simulations with an alphabet of 100 symbols: 34 were assigned a neutral label, 33 a positive label, and 33 a negative one.

As a first step we generated the {\em training sets}\footnote{In fact in our simulations there is no training, given that the classifiers operate according to a kind of Nearest Neighbor (NN) strategy, based on IoU similarities.}  used by the three base classifiers. Let us indicate these sets by ${\mathcal A}$, ${\mathcal B}$ and ${\mathcal C}$, and the corresponding classifiers by $\phi_{\mathcal A}$, $\phi_{\mathcal B}$ and $\phi_{\mathcal C}$.
All the images (multisets) composing the original datasets were generated independently. A multiset is generated by drawing symbols at random from the alphabet. Image labels are assigned using the majority-based criterion shown in \eqref{eq:majority_rule}. We run several simulations considering different multiset cardinalities, to evaluate the effect that the complexity (here approximated by the number of symbols contained in a multiset) of the images has on the effectiveness of Selecta. For each original dataset we generated 100 multisets with positive labels, 100 with negative labels and 100 with neutral labels. We did so by a trial and error procedure - that is by generating the multisets at random until we found the desired numbers of multisets of the various classes.   

Then, we created the Selecta dataset, hereafter denoted as \( \mathcal{S} \). This was done by applying the filtering procedure described in the previous section. Specifically, we randomly generated new multisets following the same procedure used for constructing the datasets \( \mathcal{A} \), \( \mathcal{B} \), and \( \mathcal{C} \). Each multiset was classified using \( \phi_{\mathcal{A}} \), \( \phi_{\mathcal{B}} \), and \( \phi_{\mathcal{C}} \), retaining only those for which all three classifiers produced the same output.
Note that the labels of the multisets in \( \mathcal{S} \) were assigned based on the unanimous predictions of \( \phi_{\mathcal{A}} \), \( \phi_{\mathcal{B}} \), and \( \phi_{\mathcal{C}} \), rather than being computed by relying on the emotional labels of the symbols composing the multisets\footnote{This strategy mimics a real-world scenario in which the images Selecta dataset consists of are collected from the internet and labeled automatically.}.
As for the size of \( \mathcal{S} \), it contains 600 multisets per class.  

The total number of multisets in Selecta dataset is larger than that of the datasets used to train \( \phi_{\mathcal{A}} \), \( \phi_{\mathcal{B}} \), and \( \phi_{\mathcal{C}} \), since one of the main goals of Selecta is the automatic generation of a large dataset starting from small ones.

As a last step we assessed the performance of the classifier based on Selecta (say, $\phi_{\mathcal{S}}$), and compared them with the performance of \( \phi_{\mathcal{A}} \), \( \phi_{\mathcal{B}} \), and \( \phi_{\mathcal{C}} \). We also built a new classifier obtained by merging the original datasets ${\mathcal A}$, ${\mathcal B}$ and ${\mathcal C}$ (in the following we indicate such a classifier as $\phi_{\mathcal A,B,C}$). The tests were carried out by generating 10.000 multisets per class
%(labeling them by using the emotional labels of the symbols they contain)
and classifying them with the 5  available classifiers. 
The simulations were repeated using multisets of different lengths: one with 13, 14, and 15 symbols; another with 15, 16, and 17; a third with 19, 20, and 21; and finally one with 29, 30, and 31 symbols.

The results obtained by running 20 complete simulations and averaging the outcomes are reported in Table \ref{tab:simulation1_100}.

\begin{table}[ht]
\centering
\scriptsize
\caption{Simulation 1 (alphabet with 100 symbols) – Accuracies of the various classifiers on multisets of different lengths.}
\begin{tabular}{p{2.6cm}ccccc}
\hline
\textbf{Multiset Length} & $\phi_{\mathcal{A}}$ & $\phi_{\mathcal{B}}$ & $\phi_{\mathcal{C}}$ & $\phi_{\mathcal{A,B,C}}$ & $\phi_{\mathcal{S}}$ \\
\hline

13, 14, 15         & 46.70\% & 46.00\% & 46.15\% & 48.65\% & 46.96\% \\
15, 16, 17         & 47.00\% & 46.92\% & 46.36\% & 48.90\% & 46.86\% \\
19, 20, 21         & 45.53\% & 45.30\% & 45.42\% & 46.53\% & 45.63\% \\
29, 30, 31         & 44.55\% & 44.16\% & 44.45\% & 46.13\% & 45.74\% \\

\hline
\end{tabular}
\label{tab:simulation1_100}
\end{table}

The simulations show that the classifier based on Selecta performs similarly to the baseline classifiers, but performs slightly worse than a classifier based on their union. The overall performance of all classifiers in this scenario remains low. A  possible explanation for these poor performance is that the random association of symbols to multisets does not reflect the way images are actually created in real life, where visual elements are not combined randomly but follow compositional mechanisms related to meaning-making. Moreover, in a real-world scenario, it is very likely that Selecta includes new elements that were not present in the original datasets. Indeed, in our theoretical assumptions, the main advantage of Selecta lies in its ability to exploit the mechanisms of isotopy to include more varied isotopic combinations within the augmented dataset.  For these reasons, we carried out a second simulation, with the aim of modeling a scenario that takes into account the principles governing the construction of visual content.

\subsection{Second simulation}

The second simulation was designed to highlight how the presence of isotopies permits to bring new knowledge (in terms of patterns that were not present in the original dataset) into the Selecta dataset.  

To start with, now some symbols may appear in the multisets used to construct \( \mathcal{S} \) that were not present in \( \mathcal{A} \), \( \mathcal{B} \), or \( \mathcal{C} \). Moreover, the final test set includes additional symbols that are absent not only from \( \mathcal{A} \), \( \mathcal{B} \), and \( \mathcal{C} \), but also from \( \mathcal{S} \). This setup mirrors real-world scenarios in which a classifier may encounter previously unseen visual lexemes. Specifically, we randomly partitioned the alphabet into three groups: 40\% of the symbols were used to construct the original datasets \( \mathcal{A} \), \( \mathcal{B} \), and \( \mathcal{C} \); an additional 30\% was used to build Selecta; and the test set encompassed the entire set of available symbols.

A second modification was designed to introduce a simplified version of the isotopy mechanism, whereby the creators of visual content tend to include elements that align with the intended emotional polarity of the image \cite{barthes1964}\cite{floch2000visual}.

To do so, if a multiset already contains emotionally concordant symbols, additional symbols with the same emotional polarity are more likely to be added. Specifically, during the random construction of the multisets, the polarity  of the first 10 symbols is checked, using the same method applied to the full multisets. At this point, the multiset is further completed by adding two elements with the same detected polarity to the multiset, while the remaining elements are sampled as usual. This mechanism serves as a minimal approximation of a visual creator's tendency toward emotional coherence. All the rest remained as in the first simulation. 

\begin{table}[ht]
\centering
\scriptsize
\caption{Simulation 2 – Results for 100-symbol configuration across different multiset lengths.}
\begin{tabular}{p{2.6cm}ccccc}
\hline
\textbf{Multiset Length} & $\phi_{\mathcal{A}}$ & $\phi_{\mathcal{B}}$ & $\phi_{\mathcal{C}}$ & $\phi_{\mathcal{A,B,C}}$ & $\phi_{\mathcal{S}}$ \\
\hline
13, 14, 15         & 48.87\% & 50.04\% & 49.30\% & 51.08\% & 56.12\% \\
15, 16, 17         & 49.29\% & 49.92\% & 49.88\% & 51.39\% & 58.01\% \\
19, 20, 21         & 49.80\% & 48.68\% & 49.84\% & 50.57\% & 56.52\% \\
29, 30, 31         & 49.91\% & 48.40\% & 48.98\% & 50.37\% & 57.57\% \\
\hline
\end{tabular}
\label{tab:simulation2_100}
\end{table}

\begin{table}[ht]
\centering
\scriptsize
\caption{Simulation statistics – Presence of patterns A and B in the images classified correctly by the different classifiers}
\begin{tabular}{p{2.8cm}ccccc}
\hline
\textbf{Pattern type} & $\phi_{\mathcal{A}}$ & $\phi_{\mathcal{B}}$ & $\phi_{\mathcal{C}}$ & $\phi_{\mathcal{A,B,C}}$ & $\phi_{\mathcal{S}}$ \\
\hline
\multicolumn{6}{c}{\textbf{Simulation 1}} \\
\hline
Pattern A observed    & 9612 & 9092 & 8468 & 27172 & 81715 \\
Common pattern B      & & & & & \\
\quad Triplets        & 0.42\% & 0.28\% & 0.39\% & 3.26\% & 17.20\% \\
\quad Quadruplets     & 0.00\% & 0.00\% & 0.00\% & 0.01\% & 0.04\% \\
\hline
\multicolumn{6}{c}{\textbf{Simulation 2}} \\
\hline
Pattern A observed    & 6365 & 6699 & 7569 & 20633 & 235969 \\
Common pattern B   & & & & & \\
\quad Triplets        & 8.20\% & 7.84\% & 7.63\% & 36.66\% & 58.17\% \\
\quad Quadruplets     & 0.17\% & 0.14\% & 0.07\% & 0.73\% & 3.97\% \\
\hline
\end{tabular}
\label{tab:simulation_statistics}
\end{table}

The results of the second group of simulations are shown in Table \ref{tab:simulation2_100}. Each simulation was repeated twenty times, and the reported values represent the average across all 20 runs. 

In the new setting, the presence of the isotopy mechanism leads to improved performance across all classifiers. Most notably, Selecta now significantly outperforms both the individual base classifiers and the classifier based on the union of the three original datasets. 
Simulation results also show that the superior performance of Selecta is not due to the size of the dataset. In the first simulation, in fact, Selecta's performance is lower than that obtained by relying on the union of the three initial datasets. In the second simulation, however, performance is much better, despite the size of the dataset being the same.
As discussed later, the good performance instead appear to be related to the number of combinations contained in Selecta dataset and to the more balanced distribution of patterns, which is particularly good in the second simulation. 

The simulations allow us to verify the hypotheses behind our methodology, which are difficult to validate on real images due to the complexity of real data analysis.

First, we hypothesized that the construction of Selecta would enable the dataset to capture numerous emotionally determinant combinations (Pattern A). In the simulations, Pattern A corresponds to symbol combinations that define the emotional content of an image - achieving an absolute majority for neutral images, or a relative majority for positive and negative ones. Every correctly classified image must contain at least one such combination, and may include several. All subsets of concordant elements sufficient to establish the majority are considered Pattern A. Table~\ref{tab:simulation_statistics} reports the number of these combinations for each classifier - that is, the emotionally relevant and sufficient combinations found among images correctly assigned to the training classes. The results in Table~\ref{tab:simulation_statistics} refer to strings of length 13, 14, and 15 symbols.

The results show that Selecta generates a dataset with a significantly larger number of pattern A. In fact, the dataset produced with Selecta is twice the size of the three datasets combined, and the number of pattern A actually observed is considerably higher, even when normalized by dataset size. This difference is  more pronounced in simulation 2 when the mechanism of isotopy is simulated more precisely.

We also argued that Selecta should promote a more balanced distribution of Pattern B across all the classes, to prevent that their presence introduces a classification bias.
%These are more frequently present in one type of emotional image, without being sufficient on their own to determine the emotion, and appear, albeit less often, in other classes, thus potentially introducing bias.
%A balanced distribution across classes can reduce such biases.
In our simulation we identify type B patterns as triplets and quartets of symbols belonging to the same emotional class. In fact, triplets and quadruplets have an impact on classification equivalent to that of pattern B. Although they are not sufficient on their own to determine the emotion of an image, in some cases they are long enough to influence the classification based on IoU and may introduce classification errors and bias. For this reason, we verified whether the dataset constructed with Selecta favors the distribution of these patterns more evenly across all classes. Table~\ref{tab:simulation_statistics} reports the percentage of triplets and quadruplets present in all classes. Specifically, the table shows the percentage of concordant triplets and quadruplets shared across all three classes, with respect to the total number of all concordant triplets and quadruplets present in the multi-sets of the training set. The results show that, in the Selecta dataset, a higher proportion of the same triplets and, to a lesser extent, quadruplets appear in all three classes rather than in only one or two, confirming a more balanced overall distribution of pattern B.

Finally, comparing the first and the second simulation, we can conclude that the good performances of Selecta are indeed linked to the exploitation of the semiotic mechanism. In fact, in the second simulation, the performance of the classifiers based on Selecta are significantly superior to both those of the individual classifier and those of the classifier based on the union of the original datasets, while this was not the case in the first simulation.

\section{Creation of an isotopy-based dataset}
\label{sec.actualdataset}

To definitively validate the Selecta methodology, we applied it to the creation of a large tripolar image dataset, starting from three smaller datasets. We then trained a VSA classifier on the new dataset and verified that it outperforms the classifiers trained individually on each of the three original datasets, as well as a classifier trained on their union. Our experiments were conducted using three existing VSA datasets: VippSent~\cite{VippSent_2025_WACV}, Flickr~\cite{katsurai2016}, and Instagram~\cite{katsurai2016}, each organized into positive, negative, and neutral classes.
%These datasets are among the most reliable for training algorithms that achieve high classification accuracy, particularly within datasets that include a neutral class.

We trained three architectures on each of the three datasets: a CLIP Image Encoder~\cite{radford2021clip} followed by a multilayer perceptron (MLP), ResNeXt101 32x8d~\cite{xie2017resnext} pretrained on ImageNet, and Swin-B~\cite{liu2021swin} pretrained on ImageNet-22k. Among these, we selected CLIP+MLP for subsequent experiments, as it achieved superior performance across all datasets (see Table \ref{tab:merged}). The three CLIP-based models, each trained on one of the original datasets, were finally used as filters to select images based on the emotions they evoke.

\begin{figure*}[htbp]
    \centering
    \includegraphics[width=\linewidth]{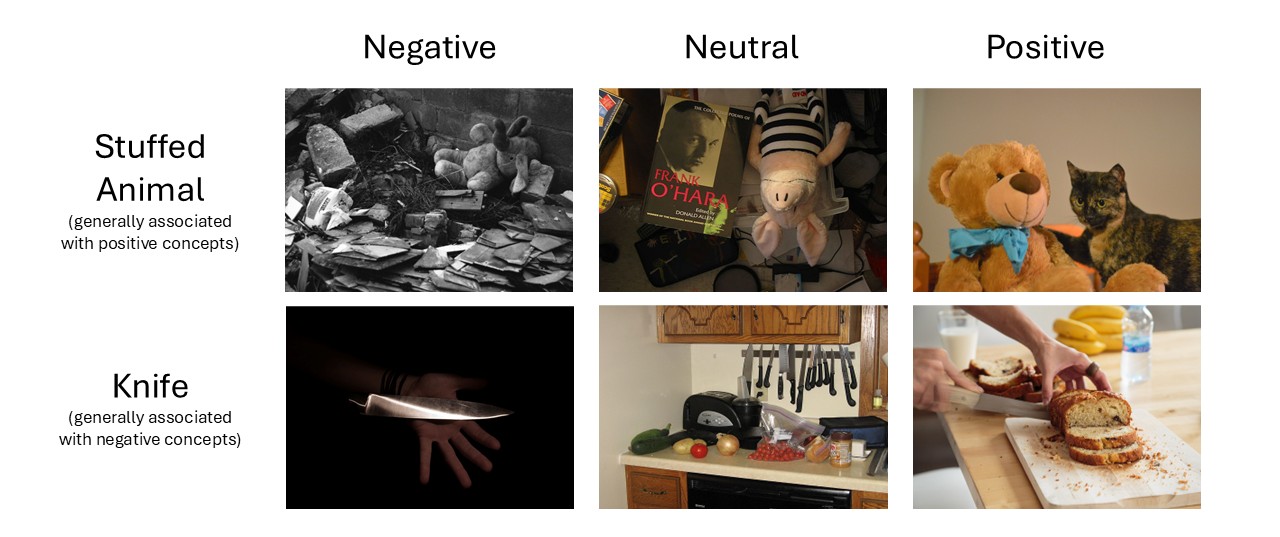}
    \caption{Examples of images with combinations of elements not present in the original datasets. In particular, a knife (typically associated with negative content) and a stuffed animal (typically associated with positive content) appear in all three classes, preventing classification from being determined solely by their presence regardless of context.}
    \label{fig:selecta_images}
\end{figure*}

To build the Selecta dataset, we downloaded two million images from the ImageNet~\cite{imagenet}, COCO~\cite{cocolin2014microsoft}, and OpenImages~\cite{openimages2020open} datasets, which were used as the image pool to be filtered. These datasets include a wide variety of images, including social media content, artistic imagery, and journalistic photography, allowing us to cover a broad spectrum of visual cases.
We then labeled the images using the three base models. Regarding the consensus strategy, only the images classified in the same way by the three models were retained and labeled accordingly (3-out-of-3 consensus).
The result was a large dataset of approximately 400,000 images. Images in the positive class have a high probability of containing positive visual isotopies, those in the negative class are likely to contain negative visual isotopies, while those in the neutral class are characterized by the absence of emotionally salient visual patterns. In this way, we built a very rich dataset, maintaining label reliability and introducing new information through the principle of isotopy. Figure \ref{fig:selecta_images} shows three images from the Selecta dataset that contain combinations of elements not present in the three original datasets. The absence of these combinations in the original datasets was verified using the YOLO object-detection software \cite{yolo8improved_yolov8_small_objects}. The examples show the images of two elements, one often associated with negative emotions, a {\em knife}, and one usually associated with positive emotions, a {\em stuffed animal}. These two elements are distributed across all three classes in Selecta, reducing the likelihood of potential biases.

Finally, we compared the performance of the models trained on the new dataset with those trained on the three original datasets and on their union. All tests were conducted on datasets used in psychological studies, due to the high reliability of their annotations.

For training, the datasets were split into training, validation, and test sets using a 70:15:15 ratio. To increase variability during training, we applied data augmentation techniques such as JPEG compression with random quality between 60 and 95, and horizontal and vertical flips, while avoiding color transformations in order to preserve the sentiment conveyed by the images.
The models were trained for up to 100 epochs, selecting the best-performing version. We used the Adam optimizer with an initial learning rate of 0.0001, adjusted using a linear scheduler, and a batch size of 64.
Images were resized to 336 $\times$ 336. These settings were maintained for all the experiments, except for those performed on networks other than CLIP, for which the size was set to 224 $\times$ 224.

\section{Experimental validation}
\label{sec:experiment}

In this section, we compare the performance of models trained on Selecta with those trained on the original datasets and their union. We evaluate the performance on both tripolar and bipolar datasets.

\subsection{Comparison with base classifiers}

For our tests, we trained three models: CLIP \cite{radford2021clip} with an additional MLP classifier, ResNet-101 \cite{he2016resnet}, and Swin-T \cite{liu2021swin}. All models were trained both on Selecta and on the three original datasets (VippSent, Flickr, and Instagram).

Cross testing was conducted on eight of the most widely used datasets in the literature: four psychological datasets with a neutral class (IAPS~\cite{Iaps}, GAPED~\cite{gaped}, OASIS~\cite{OasisKurdiLozanoBanaji2017}, NAPS~\cite{NapsMarchewkaZurawskiJednorogGrabowska2014}) and four VSA benchmarks including only two classes (FI~\cite{FIYou2016BuildingAL}, Emotion6~\cite{Emotion6}, Emotic~\cite{emotic}, VSO~\cite{VSOBorth2013}).
Differently from the comparison shown in Table I, in this experiment on the two-class datasets we applied the models without any modification, leaving the neutral class empty in the test with the expected outcome that no images would be assigned to it. This allowed us to also assess the effectiveness of the models in the classification of neutral images, even in the two-class test (positive and negative).

\renewcommand{\arraystretch}{1.8} 
\begin{table*}[h!]
\centering
\caption{Accuracies of the models on various test sets. For each architecture (CLIP, Swin-B, ResNet101), the results are reported for models trained on the VIPP, Flickr, Instagram, Union, and Selecta datasets.}
\label{tab:merged}

\large

\resizebox{\textwidth}{!}{%
\begin{tabular}{|l|c|c|c|c|c|c|c|c|c|c|c|c|c|c|c|}
\hline
\textbf{Dataset} 
& \multicolumn{5}{c|}{\textbf{CLIP}} 
& \multicolumn{5}{c|}{\textbf{Swin-B}} 
& \multicolumn{5}{c|}{\textbf{ResNet101}} \\ \hline
& \textbf{VIPP} & \textbf{Flk.} & \textbf{Insta} & \textbf{Un.} & \underline{\textbf{Sel.}}
& \textbf{VIPP} & \textbf{Flk.} & \textbf{Insta} & \textbf{Un.} & \underline{\textbf{Sel.}}
& \textbf{VIPP} & \textbf{Flk.} & \textbf{Insta} & \textbf{Un.} & \underline{\textbf{Sel.}} \\ \hline
Own Test-Set 
& 88.64\% & 75.71\% & 70.45\% & 76.26\% & \underline{96.04\%}
& 85.07\% & 72.91\% & 66.42\% & 73.03\% & \underline{87.78\%}
& 83.16\% & 71.07\% & 64.53\% & 71.92\% & \underline{86.12\%} \\ \hline
IAPS 3 cl. 
& 84.41\% & 75.15\% & 78.17\% & 84.23\% & \underline{89.34\%}
& 81.38\% & 72.41\% & 74.91\% & 64.78\% & \underline{82.14\%}
& 79.36\% & 70.45\% & 73.12\% & 61.71\% & \underline{80.82\%} \\ \hline
GAPED 3 cl. 
& 83.53\% & 76.08\% & 77.22\% & 84.15\% & \underline{88.52\%}
& 79.29\% & 72.78\% & 73.91\% & 68.93\% & \underline{85.74\%}
& 77.11\% & 70.19\% & 72.01\% & 65.92\% & \underline{79.06\%} \\ \hline
NAPS 3 cl. 
& 60.29\% & 51.10\% & 61.10\% & 60.74\% & \underline{68.61\%}
& 56.71\% & 48.86\% & 57.05\% & 58.81\% & \underline{67.27\%}
& 54.62\% & 46.87\% & 55.28\% & 53.69\% & \underline{60.99\%} \\ \hline
OASIS 3 cl. 
& 70.09\% & 63.34\% & 76.66\% & 73.64\% & \underline{78.65\%}
& 66.89\% & 60.77\% & 72.15\% & 64.49\% & \underline{67.36\%}
& 64.01\% & 58.04\% & 70.26\% & 64.72\% & \underline{68.73\%} \\ \hline
FI 2 cl. 
& 80.11\% & 49.00\% & 80.13\% & 77.66\% & \underline{82.91\%}
& 76.28\% & 46.77\% & 76.49\% & 70.18\% & \underline{77.97\%}
& 73.91\% & 43.69\% & 74.11\% & 70.31\% & \underline{76.69\%} \\ \hline
VSO 2 cl. 
& 56.61\% & 40.25\% & 53.35\% & 49.68\% & \underline{56.82\%}
& 52.74\% & 37.11\% & 50.19\% & 52.05\% & \underline{53.91\%}
& 50.11\% & 34.88\% & 47.33\% & 49.74\% & \underline{51.73\%} \\ \hline
Emot. 6 2 cl. 
& 80.79\% & 58.93\% & 70.06\% & 75.92\% & \underline{81.85\%}
& 76.92\% & 55.02\% & 67.44\% & 67.77\% & \underline{78.63\%}
& 75.11\% & 52.08\% & 64.73\% & 66.94\% & \underline{77.84\%} \\ \hline
Emotic 2 cl. 
& 79.00\% & 60.50\% & 69.00\% & 79.32\% & \underline{82.45\%}
& 75.16\% & 56.12\% & 65.74\% & 63.69\% & \underline{77.85\%}
& 73.19\% & 53.22\% & 63.12\% & 58.71\% & \underline{78.94\%} \\ \hline
\end{tabular}%
}
\end{table*}

As it can be seen from Table~\ref{tab:merged}, the model trained on Selecta consistently outperforms the other models, with gains ranging from 1\% to 8\%.
In the table, the Own Test Set row reports the performance obtained on the test set of the dataset used for training, while the other rows report cross-testing results. The table reports results obtained by training the models on three classes. For tests on 2-class datasets, images assigned to the neutral class are considered as errors.

We further compared the results of models trained on Selecta with those trained on the union of the Flickr, VippSent, and Instagram datasets.
The goal is to verify whether, even when merging the three original datasets, Selecta is able to maintain superior generalization capabilities.
Similarly, we trained three architectures (CLIP, Swin Transformer, and ResNet-101) both on Selecta and on the union of the original datasets, comparing their performance on the main Visual Sentiment Analysis datasets, with either two classes (positive, negative) or three classes (positive, negative, neutral).
Since the models were designed for three classes, in this case as well an empty neutral class was added in the tests on two-class datasets to enable evaluation.

The results we obtained, shown in Table~\ref{tab:merged}, indicate that Selecta consistently achieves better performance across all test sets, in line with the trends predicted by the simulations.

\subsection{Comparison of results across different emotional classes}

As a further experiment, we compared the performance of classifiers trained on Selecta with those trained on the union of the three original datasets across the three emotional classes. We focused on CLIP models, given their significantly superior performance. These models were tested on IAPS and GAPED, selected for the reliability of their labels, with the aim of assessing how the improvements obtained with Selecta are distributed across the different emotional classes.

The results in Table \ref{tab:differenti_classi} show a clear improvement in the neutral class. This observation is consistent both with the predictions of the isotopy mechanism theorized by semiotics and with the considerations outlined in the methodology. Positive and negative images, in fact, contain recurring visual elements that enable good performance in both cases. Neutrality, on the other hand, is characterized by more heterogeneous elements and by a lower recurrence of typical visual patterns, since it emerges from the absence of emotional isotopies. For this reason, the neutral class is more difficult to classify: in fact, the lack of typical visual forms increases the likelihood of biases caused by Patterns C and B.

The model trained on Selecta, having been exposed to a wider variety of isotopic combinations and to a better distribution of Pattern B across the three classes, proves to be more resistant to these classification biases. For this reason, it distinguishes the neutral class better than models trained on the original datasets.

\begin{table}[h!]
\centering
\caption{Results of CLIP-based models trained on Selecta and the union dataset, on IAPS and GAPED split by emotional class}
\label{tab:differenti_classi}
\renewcommand{\arraystretch}{1.4}
\resizebox{0.5\textwidth}{!}{%
\begin{tabular}{|l|c|c|c|c|c|c|}
\hline
\textbf{Model} 
& \shortstack{\textbf{Neg}\\\textbf{IAPS}} 
& \shortstack{\textbf{Neu}\\\textbf{IAPS}} 
& \shortstack{\textbf{Pos}\\\textbf{IAPS}} 
& \shortstack{\textbf{Neg}\\\textbf{GAPED}} 
& \shortstack{\textbf{Neu}\\\textbf{GAPED}} 
& \shortstack{\textbf{Pos}\\\textbf{GAPED}} \\ \hline
\textbf{Selecta CLIP} & 94.06\% & 71.32\% & 86.76\% & 96.11\% & 80.85\% & 85.44\% \\ \hline
\textbf{Union CLIP}   & 91.66\% & 55.52\% & 83.21\% & 94.72\% & 65.12\% & 82.37\% \\ \hline
\end{tabular}
}
\end{table}

\subsection{Two-class experiments}

As a further validation, we compared Selecta with Emoset (containing only positive and negative images), to verify whether Selecta's generalization capabilities are confirmed even in the absence of the neutral class.
To this end, we trained the models both on Selecta and Emoset by excluding the neutral class. As before, we trained three architectures: CLIP+MLP, Swin-B, and ResNet-101. Tests were carried out on the positive and negative images of the same datasets used in the previous experiments. 
\renewcommand{\arraystretch}{1.4} 
\begin{table}[h!]
\centering
\caption{Accuracies of models trained on the positive and negative classes of Selecta and Emoset.}
\label{table3}
\small % <-- oppure \footnotesize, a seconda di quanto vuoi ridurre
\resizebox{0.5\textwidth}{!}{%
\begin{tabular}{|l|c|c|c|c|c|c|}
\hline
\textbf{Dataset} & \shortstack{\textbf{Sel.}\\\textbf{Clip}} & \shortstack{\textbf{Emo.}\\\textbf{Clip}} & \shortstack{\textbf{Sel.}\\\textbf{SwinB}} & \shortstack{\textbf{Emo.}\\\textbf{SwinB}} & \shortstack{\textbf{Sel.}\\\textbf{Res101}} & \shortstack{\textbf{Emo.}\\\textbf{Res101}} \\ \hline
Own Test Set & 99.21\% & 94.68\% & 91.74\% & 90.65\% & 92.83\% & 92.79\% \\ \hline
IAPS 2 cl. & 97.22\% & 93.76\% & 79.94\% & 77.53\% & 78.72\% & 76.21\% \\ \hline
GAPED 2 cl. & 93.09\% & 91.19\% & 85.63\% & 81.71\% & 84.82\% & 80.94\% \\ \hline
NAPS 2 cl. & 75.44\% & 73.82\% & 72.73\% & 70.69\% & 71.84\% & 70.83\% \\ \hline
OASIS 2 cl. & 92.88\% & 86.92\% & 80.68\% & 78.72\% & 78.69\% & 76.63\% \\ \hline
FI & 88.64\% & 86.30\% & 72.64\% & 69.91\% & 71.79\% & 67.91\% \\ \hline
VSO & 66.81\% & 65.75\% & 65.74\% & 58.91\% & 63.92\% & 55.82\% \\ \hline
Emotion6 & 86.40\% & 83.93\% & 71.94\% & 69.77\% & 70.68\% & 68.71\% \\ \hline
Emotic & 91.95\% & 87.63\% & 71.72\% & 68.74\% & 71.65\% & 67.89\% \\ \hline
\end{tabular}%
}
\end{table}

The results we got are shown in Table~\ref{table3}. Even in this case, models trained on Selecta achieve superior performance across all test sets, further proving the validity of our approach.

\subsection{3-out-of-3 vs 2-out-of-3 consensus}

The last test we run aimed at comparing two different consensus strategies: a more selective one, based on full agreement among the three base algorithms, and a looser one, for which an agreement is reached when at least two out of three based classifiers give the same result. In the first case, the selection process filtered 400,000 images, 600,000 in the second case.

\begin{table}[h!]
\centering
\caption{
Comparison of different consensus strategies.}
\label{tab:selecta_comparison}
\renewcommand{\arraystretch}{1.4}
\begin{tabular}{|l|c|c|c|c|c|}
\hline
\textbf{Model} & \textbf{Test} & \textbf{IAPS} & \textbf{GAPED} & \textbf{NAPS} & \textbf{OASIS} \\ \hline
\textbf{Selecta 3/3 Clip} & 96.04\% & 89.34\% & 88.52\% & 68.61\% & 78.65\% \\ \hline
\textbf{Selecta 2/3 Clip} & 90.78\% & 85.69\% & 85.94\% & 61.73\% & 75.84\% \\ \hline
\end{tabular}
\end{table}

Upon inspection of the results in Table~\ref{tab:selecta_comparison}, we see that despite the smaller number of images, the dataset built on a stricter consensus strategy achieves better performance across all four psychological datasets. This result suggests that the selection of images based on isotopy mechanisms has a greater impact on model performance than dataset size alone.

\section{Conclusion}
\label{sec:conclusion}

We proposed a new methodology for constructing a VSA dataset with improved generalization capabilities.
The methodology is grounded on the analysis of the typical challenges in VSA, particularly the difficulty of classifying images according to abstract labels such as emotions, which lack univocal visual correspondences. To address this issue, we drew on concepts from semiotics, including isotopy and emotional signification, and introduced a dataset design strategy enriched with emotionally relevant visual combinations. This research shows that the theoretical tools developed within the field of visual semiotics can be employed to improve performance in image classification, particularly in complex tasks such as emotion classification. Moreover, it demonstrates that a careful and targeted data selection procedure in VSA, in our case backed by semiotic principles, proved to be a key factor in significantly enhancing model performance, as also highlighted in other areas of visual computing \cite{geirhos2019imagenet}.

\subsection{Limitations and Future Work}

A  limitation of the present work concerns the use of only three emotional classes: positive, negative, and neutral. This choice was adopted for both practical and theoretical reasons.

From a practical perspective, there is a need to include a neutral class, which is a fundamental element for real-world VSA applications, in order to avoid forcing images that do not evoke emotions into an emotional category, and a sufficiently broad and accurate neutral class is available only in three-class datasets. A further constraint is represented by the availability of reference datasets: to apply our methodology, at least three reliable datasets classified according to the same emotional classes are required. Datasets with a larger number of categories are often too small or based on heterogeneous annotation schemes, such as Ekman’s six basic emotions~\cite{Ekman1992}, Mikels’ model~\cite{mikels2005emotional}, or other classifications.

From a theoretical perspective, classifications with a larger number of categories are generally limited to the basic emotions of Ekman and Mikels, while studies in psychology show that images can also evoke more complex feelings, such as jealousy or nostalgia, which remain excluded from such schemes~\cite{Ekman1992, mikels2005emotional, panksepp1998affective, Damasio2003, Barrett2017ConstructedEmotion}. The three-class structure, already adopted in psychology~\cite{Russell1980}, better reflects this complexity by including both basic emotions and more articulated feelings. Moreover, it aligns with the categories used in semiotics to analyze the affective aspects of a text: euphoria (positive), dysphoria (negative), and aphoria (neutral)\cite{greimas1993semiotics}. Having said that, however, the Selecta methodology can also be applied to datasets with different emotional classifications, such as Ekman's and Mikels', provided that a sufficient number of reliable datasets based on the same type of emotional classes are available.

Another promising direction for future research concerns the analysis of multimodal content, particularly the integrated composition of images and text. Extending the principle of isotopy to both modalities could enable the application of Selecta to textual content analysis as well.

Finally, there is a broader limitation that affects the entire VSA field. Emotional responses to images are shaped by individual, cultural, and contextual factors. Moreover, many images are inherently ambiguous or may acquire different emotional meanings depending on the context in which they are viewed~\cite{barthes1964}. Consequently, visual sentiment analysis typically seeks to identify the emotion perceived by the majority of observers, while acknowledging that such responses are not universal. A potential direction for future work would be the development of a dataset that explicitly includes a class for emotionally ambiguous images.

\bibliographystyle{IEEEtran}
\bibliography{bibliografia}

@String(CVPR= {IEEE Conf. Comput. Vis. Pattern Recog.})

@String(ECCV= {Eur. Conf. Comput. Vis.})

@String(ICASSP=	{ICASSP})

@String(ICLR = {Int. Conf. Learn. Represent.})

@String(AAAI = {AAAI})

@String(CVPRW= {IEEE Conf. Comput. Vis. Pattern Recog. Worksh.})

@String(CVPR  = {CVPR})

@String(ECCV  = {ECCV})

@String(ICLR  = {ICLR})

@String(CVPRW= {CVPRW})

@article{Ortis2020,
  author    = {Ortis, Alessandro and Farinella, Giovanni Maria and Battiato, Sebastiano},
  title     = {\textit{Survey on visual sentiment analysis}},
  journal   = {IET Image Processing},
  volume    = {14},
  number    = {8},
  pages     = {1440--1456},
  year      = {2020}
}

@article{mikels2005emotional,
  author    = {Mikels, Joseph A and Fredrickson, Barbara L and Larkin, Gregory R and Lindberg, Casey M and Maglio, Sam J and Reuter-Lorenz, Patricia A},
  title     = {\textit{Emotional Category Data on Images from the International Affective Picture System}},
  journal   = {Behavior Research Methods},
  volume    = {37},
  number    = {4},
  pages     = {626--630},
  year      = {2005}
}

@inproceedings{katsurai2016,
  author    = {Katsurai, Marie and Satoh, Shin'ichi},
  title     = {\textit{Image sentiment analysis using latent correlations among visual, textual, and sentiment views}},
  booktitle = {2016 IEEE International Conference on Acoustics, Speech and Signal Processing (ICASSP)},
  pages     = {2837--2841},
  year      = {2016}
}

@article{yang2023L,
  author    = {Yang, Jingyuan and Huang, Qirui and Ding, Tingting and Lischinski, Dani and Cohen-Or, Daniel and Huang, Hui},
  title     = {\textit{EmoSet: A Large-scale Visual Emotion Dataset with Rich Attributes}},
  journal   = {Proceedings of the IEEE/CVF International Conference on Computer Vision},
  pages     = {20326--20337},
  year      = {2023}
}

@article{barthes1964,
  author    = {Barthes, Roland},
  title     = {Rhetoric of the Image},
  booktitle = {Visual Culture: The Reader},
  editor    = {Evans, Jessica and Hall, Stuart},
  publisher = {Sage Publications},
  address   = {London},
  pages     = {33--40},
  year      = {1999},
  note      = {Originally published in 1964 in \textit{Communications}}
}

@inproceedings{he2016resnet,
  author    = {He, Kaiming and Zhang, Xiangyu and Ren, Shaoqing and Sun, Jian},
  title     = {\textit{Deep Residual Learning for Image Recognition}},
  booktitle = {Proceedings of the IEEE Conference on Computer Vision and Pattern Recognition},
  pages     = {770--778},
  publisher = {IEEE Computer Society},
  year      = {2016}
}

@inproceedings{xie2017resnext,
  author    = {Xie, Saining and Girshick, Ross B. and Doll{\'a}r, Piotr and Tu, Zhuowen and He, Kaiming},
  title     = {\textit{Aggregated Residual Transformations for Deep Neural Networks}},
  booktitle = {Proceedings of the IEEE Conference on Computer Vision and Pattern Recognition},
  pages     = {5987--5995},
  publisher = {IEEE Computer Society},
  year      = {2017}
}

@inproceedings{liu2021swin,
  author    = {Liu, Ze and Lin, Yutong and Cao, Yue and Hu, Han and Wei, Yixuan and Zhang, Zheng and Lin, Stephen and Guo, Baining},
  title     = {\textit{Swin Transformer: Hierarchical Vision Transformer using Shifted Windows}},
  booktitle = {Proceedings of the IEEE/CVF International Conference on Computer Vision},
  pages     = {9992--10002},
  publisher = {IEEE},
  year      = {2021}
}

@inproceedings{radford2021clip,
  author    = {Radford, Alec and Kim, Jong Wook and Hallacy, Chris and Ramesh, Aditya and Goh, Gabriel and Agarwal, Sandhini and Sastry, Girish and Askell, Amanda and Mishkin, Pamela and Clark, Jack and Krueger, Gretchen and Sutskever, Ilya},
  title     = {\textit{Learning Transferable Visual Models From Natural Language Supervision}},
  booktitle = {Proceedings of the 38th International Conference on Machine Learning},
  volume    = {139},
  pages     = {8748--8763},
  year      = {2021}
}

@inproceedings{Emotion6,
  author    = {Peng, Kuan-Chuan and Chen, Tsuhan and Sadovnik, Amir and Gallagher, Andrew C.},
  title     = {A mixed bag of emotions: Model, predict, and transfer emotion distributions},
  booktitle = {IEEE Conference on Computer Vision and Pattern Recognition (CVPR)},
  year      = {2015},
  pages     = {860--868}
}

@inproceedings{t4s2017,
  author    = {Vadicamo, Lucia and Carrara, Fabio and Cimino, Andrea and Cresci, Stefano and Dell'Orletta, Felice and Falchi, Fabrizio and Tesconi, Maurizio},
  title     = {Cross-Media Learning for Image Sentiment Analysis in the Wild},
  booktitle = {IEEE International Conference on Computer Vision Workshops (ICCVW)},
  year      = {2017},
  pages     = {308--317},
  doi       = {10.1109/ICCVW.2017.45}
}

@article{gaped,
  author  = {Dan-Glauser, Elise S. and Scherer, Klaus R.},
  title   = {The Geneva affective picture database (GAPED): a new 730-picture database focusing on valence and normative significance},
  journal = {Behavior Research Methods},
  year    = {2011},
  volume  = {43},
  pages   = {468--477}
}

@inproceedings{emotic,
  author    = {Kosti, Ronak and Alvarez, Jose M. and Recasens, Adria and Lapedriza, {\`A}gata},
  title     = {EMOTIC: Emotions in Context Dataset},
  booktitle = {IEEE Conference on Computer Vision and Pattern Recognition Workshops (CVPRW)},
  year      = {2017},
  pages     = {2309--2317},
  doi       = {10.1109/CVPRW.2017.285}
}

@inproceedings{FIYou2016BuildingAL,
  author    = {You, Quanzeng and Luo, Jiebo and Jin, Hailin and Yang, Jianchao},
  title     = {Building a Large Scale Dataset for Image Emotion Recognition: The Fine Print and the Benchmark},
  booktitle = {Proceedings of the AAAI Conference on Artificial Intelligence (AAAI)},
  volume    = {30},
  number    = {1},
  year      = {2016},
  pages     = {308--314}
}

@book{greimas1993semiotics,
  author    = {Greimas, Algirdas Julien and Fontanille, Jacques},
  title     = {The Semiotics of Passions: From States of Affairs to States of Feeling},
  publisher = {University of Minnesota Press},
  year      = {1993}
}

@inproceedings{Iaps,
  author    = {Lang, Peter J. and Bradley, Margaret M. and Cuthbert, Bruce N.},
  title     = {International Affective Picture System (IAPS) [Database record]},
  year      = {2005},
  howpublished = {APA PsycTests},
  doi       = {10.1037/t66667-000}
}

@book{floch2000visual,
  author    = {Floch, Jean-Marie},
  title     = {Visual Identities: Constructing the Self in Western Visual Culture},
  publisher = {Bloomsbury Academic},
  address   = {London},
  year      = {2000},
  series    = {Continuum Studies in Semiotics},
  isbn      = {9780826447388},
  note      = {Translated from the French by Andrew Brown}
}

@book{greimas1987meaning,
  author    = {Greimas, Algirdas Julien},
  editor    = {Perron, Paul J. and Collins, Frank H.},
  title     = {On Meaning: Selected Writings in Semiotic Theory},
  publisher = {University of Minnesota Press},
  address   = {Minneapolis},
  year      = {1987},
  series    = {Theory and History of Literature},
  volume    = {38},
  isbn      = {9780816615186},
}

@article{dimensionalscherer2009,
  author    = {Scherer, Klaus R.},
  title     = {Dimensional and Categorical Approaches to Emotion: Theory and Research},
  journal   = {Emotion Review},
  year      = {2009},
  volume    = {1},
  number    = {3},
  pages     = {292--300},
  doi       = {10.1177/1754073909103591}
}

@article{You2015,
  author    = {You, Quanzeng and Luo, Jiebo and Jin, Hailin and Yang, Jianchao},
  title     = {Robust Image Sentiment Analysis Using Progressively Trained and Domain Transferred Deep Networks},
  booktitle = {Proceedings of the AAAI Conference on Artificial Intelligence (AAAI)},
  volume    = {29},
  number    = {1},
  year      = {2015},
  pages     = {381--388}
}

@article{Peng2016,
  author    = {Peng, Kaicheng and Zhu, Yi and Ma, Xiaoxi and Tan, Chong-Wah and Yan, Shuicheng},
  title     = {Image sentiment analysis using deep convolutional neural networks with domain transfer},
  journal   = {IEEE Transactions on Affective Computing},
  year      = {2016},
  volume    = {9},
  number    = {3},
  pages     = {352--370}
}

@inproceedings{Campos2017,
  author    = {Victor Campos and Brendan Jou and Xavier Gir{\'o}-i-Nieto},
  title     = {\emph{Sentiment analysis in social multimedia: Methodologies and applications}},
  booktitle = {Proceedings of the ACM International Conference on Multimedia Retrieval (ICMR)},
  pages     = {411--416},
  year      = {2017},
  publisher = {ACM}
}

@article{Zhou2020,
  author    = {Zixing Zhou and Yiming Yu and Ivor Wai-Hung Gu},
  title     = {\emph{A survey on visual sentiment analysis}},
  journal   = {Artificial Intelligence Review},
  volume    = {53},
  number    = {3},
  pages     = {1801--1819},
  year      = {2020},
  publisher = {Springer}
}

@book{Damasio2003,
  author    = {Antonio R. Damasio},
  title     = {\emph{Looking for Spinoza: Joy, Sorrow, and the Feeling Brain}},
  publisher = {Harcourt},
  address   = {New York},
  year      = {2003},
}

@book{panksepp1998affective,
  author    = {Jaak Panksepp},
  title     = {\emph{Affective Neuroscience: The Foundations of Human and Animal Emotions}},
  year      = {1998},
  publisher = {Oxford University Press},
  address   = {New York, NY, USA},
  isbn      = {9780195096736}
}

@article{Russell1980,
  author    = {James A. Russell},
  title     = {\emph{A circumplex model of affect}},
  journal   = {Journal of Personality and Social Psychology},
  volume    = {39},
  number    = {6},
  pages     = {1161--1178},
  year      = {1980},
  doi       = {10.1037/h0077714}
}

@inproceedings{VippSent_2025_WACV,
  author    = {Marco Blanchini and Giovanna Dimitri and Lydia Abady and Benedetta Tondi and Tarcisio Lancioni and Mauro Barni},
  title     = {\emph{Semiotic-based construction of a large emotional image dataset with neutral samples}},
  booktitle = {Proceedings of the Winter Conference on Applications of Computer Vision (WACV)},
  pages     = {7541--7550},
  month     = {February},
  year      = {2025}
}

@inproceedings{machajdik2010affective,
  author    = {Jana Machajdik and Allan Hanbury},
  title     = {\emph{Affective image classification using features inspired by psychology and art theory}},
  booktitle = {Proceedings of the International Conference on Multimedia},
  pages     = {83--92},
  year      = {2010},
  publisher = {ACM},
  doi       = {10.1145/1873951.1873965}
}

@inproceedings{geirhos2019imagenet,
  author    = {Robert Geirhos and Patricia Rubisch and Claudio Michaelis and Matthias Bethge and Felix A. Wichmann and Wieland Brendel},
  title     = {\emph{ImageNet-trained CNNs are biased towards texture; increasing shape bias improves accuracy and robustness}},
  booktitle = {International Conference on Learning Representations (ICLR)},
  year      = {2019}
}

@inproceedings{zhang2019interpreting,
  author    = {Quanshi Zhang and Yu Yang and Haotian Ma and Ying Nian Wu},
  title     = {\emph{Interpreting CNNs via Decision Trees}},
  booktitle = {Proceedings of the IEEE/CVF Conference on Computer Vision and Pattern Recognition (CVPR)},
  pages     = {6261--6270},
  year      = {2019},
  publisher = {IEEE}
}

@article{masciantonio2025positivity,
  author    = {Alexandra Masciantonio and Neele Heiser and Anthony Cherbonnier},
  title     = {\emph{Unveiling the positivity bias on social media: A registered experimental study on Facebook, Instagram, and X}},
  journal   = {Collabra: Psychology},
  volume    = {11},
  number    = {1},
  pages     = {1--17},
  year      = {2025},
  doi       = {10.1525/collabra.132410},
  publisher = {University of California Press}
}

@article{OasisKurdiLozanoBanaji2017,
  author    = {Benedek Kurdi and Shayn Lozano and Mahzarin R. Banaji},
  title     = {\emph{Introducing the Open Affective Standardized Image Set (OASIS)}},
  journal   = {Behavior Research Methods},
  volume    = {49},
  pages     = {457--470},
  year      = {2017},
  doi       = {10.3758/s13428-016-0715-3}
}

@article{NapsMarchewkaZurawskiJednorogGrabowska2014,
  author    = {Artur Marchewka and Lukasz Zurawski and Katarzyna Jednorog and Anna Grabowska},
  title     = {\emph{The Nencki Affective Picture System (NAPS): Introduction to a novel standardized wide-range high-quality realistic picture database}},
  journal   = {Behavior Research Methods},
  volume    = {46},
  number    = {2},
  pages     = {596--610},
  year      = {2014},
  doi       = {10.3758/s13428-013-0379-1}
}

@article{vuilleumier2005how,
  author    = {Patrik Vuilleumier},
  title     = {\emph{How brains beware: Neural mechanisms of emotional attention}},
  journal   = {Trends in Cognitive Sciences},
  volume    = {9},
  number    = {12},
  pages     = {585--594},
  year      = {2005},
  publisher = {Elsevier},
  doi       = {10.1016/j.tics.2005.10.011}
}

@article{Ekman1992,
  author    = {Paul Ekman},
  title     = {\emph{An argument for basic emotions}},
  journal   = {Cognition \& Emotion},
  volume    = {6},
  pages     = {169--200},
  year      = {1992}
}

@inproceedings{VSOBorth2013,
  author    = {Damian Borth and R. Ji and Tao Chen and Thomas M. Breuel and Shih-Fu Chang},
  title     = {\emph{Large-scale visual sentiment ontology and detectors using adjective noun pairs}},
  booktitle = {Proceedings of the 21st ACM International Conference on Multimedia},
  year      = {2013}
}

@book{sontag1977,
  author    = {Sontag, Susan},
  title     = {On Photography},
  publisher = {Farrar, Straus and Giroux},
  address   = {New York},
  year      = {1977},
  isbn      = {9780374224041}
}

@book{boltanski1993souffrance,
  author    = {Luc Boltanski},
  title     = {\emph{Distant Suffering: Morality, Media and Politics}},
  publisher = {Cambridge University Press},
  address   = {Cambridge},
  year      = {1999}
}

@inproceedings{imagenet,
  author    = {Jia Deng and Wei Dong and Richard Socher and Li-Jia Li and Kai Li and Fei-Fei Li},
  title     = {\emph{ImageNet: A Large-Scale Hierarchical Image Database}},
  booktitle = {2009 IEEE Conference on Computer Vision and Pattern Recognition},
  pages     = {248--255},
  year      = {2009},
  organization = {IEEE}
}

@inproceedings{cocolin2014microsoft,
  author    = {Tsung-Yi Lin and Michael Maire and Serge Belongie and James Hays and Pietro Perona and Deva Ramanan and Piotr Doll{\'a}r and C. Lawrence Zitnick},
  title     = {\emph{Microsoft COCO: Common Objects in Context}},
  booktitle = {European Conference on Computer Vision (ECCV)},
  pages     = {740--755},
  publisher = {Springer},
  year      = {2014}
}

@article{openimages2020open,
  author    = {Alina Kuznetsova and Hassan Rom and Neil Alldrin and Jasper Uijlings and Ivan Krasin and Jordi Pont-Tuset and Shahab Kamali and Stefan Popov and Matteo Malloci and Alexander Kolesnikov and Tom Duerig},
  title     = {\emph{The Open Images Dataset V4: Unified Image Classification, Object Detection, and Visual Relationship Detection at Scale}},
  journal   = {International Journal of Computer Vision},
  volume    = {128},
  pages     = {1956--1981},
  publisher = {Springer},
  year      = {2020}
}

@book{GrammaticaVisivakress2006reading,
  author    = {Gunther Kress and Theo van Leeuwen},
  title     = {\emph{Reading Images: The Grammar of Visual Design}},
  edition   = {2nd},
  publisher = {Routledge},
  address   = {London and New York},
  year      = {2006},
  isbn      = {978-0415319157}
}

@article{Barrett2017ConstructedEmotion,
  author    = {Lisa Feldman Barrett},
  title     = {\emph{The theory of constructed emotion: An active inference account of interoception and categorization}},
  journal   = {Social Cognitive and Affective Neuroscience},
  volume    = {12},
  number    = {1},
  pages     = {1--23},
  publisher = {Oxford University Press},
  year      = {2017},
  doi       = {10.1093/scan/nsw154}
}

@inproceedings{yolo8improved_yolov8_small_objects,
  author    = {Huang, H. and Zhang, Y. and Liu, X. and Wang, J. and Chen, L.},
  title     = {Improved Small-Object Detection Using YOLOv8},
  booktitle = {Proceedings of the 2023 International Conference on Machine Learning and Automation},
  year      = {2023},
  doi       = {10.54254/2755-2721/41/20230714}
}
\end{document}